\pdfoutput=1

\documentclass[11pt]{article}

\usepackage[final]{acl}

\usepackage{times}
\usepackage{latexsym}

\usepackage[T1]{fontenc}

\usepackage[utf8]{inputenc}

\usepackage{microtype}

\usepackage{inconsolata}

\usepackage{graphicx}

\usepackage{natbib}
\usepackage{booktabs}
\usepackage{multirow}
\usepackage{subcaption}
\usepackage{enumitem}
\usepackage{fdsymbol}
\usepackage{natbib}
\usepackage{CJKutf8}
\usepackage{xcolor}
\usepackage{hyperref}
\definecolor{lowblue}{hsb}{0.6, 0.3, 1}  
\definecolor{lowred}{hsb}{0, 0.3, 1}     

\title{Lost in Literalism: How Supervised Training Shapes Translationese in LLMs}

\author{ 
 Yafu Li$^{\spadesuit \clubsuit}\footnotemark[1]$\hspace{0.5mm}, 
 Ronghao Zhang$^{\clubsuit \vardiamondsuit}\footnotemark[1]$\hspace{0.5mm}, 
 Zhilin Wang$^{\spadesuit \clubsuit}$\hspace{0.5mm}, 
 Huajian Zhang$^{\clubsuit}$\hspace{0.5mm}, \\
  \bf{
  Leyang Cui$^{\clubsuit}$\hspace{0.5mm}, 
 Yongjing Yin$^{\clubsuit}$\hspace{0.5mm}, 
 Tong Xiao$^{\diamondsuit}$\hspace{0.5mm}, 
 Yue Zhang$^{\clubsuit }$\footnotemark[2]}\hspace{0.2mm}\hspace{1.5mm} \\
$^\spadesuit$ Shanghai AI Laboratory \ \ \ \quad$^\clubsuit$Westlake University \\ $^\vardiamondsuit$ Zhejiang University \ \ \ \quad$^\diamondsuit$Northeastern University\ \ \ 
 \\
 \texttt{yafuly@gmail.com} \quad \texttt{zhangyue@westlake.edu.cn} \\
}

\begin{document}
\maketitle
\renewcommand{\thefootnote}{\fnsymbol{footnote}}
\footnotetext[1]{\ Equal contributions.}
\footnotetext[2]{\ Corresponding author.}
\begin{abstract}
Large language models (LLMs) have achieved remarkable success in machine translation, demonstrating impressive performance across diverse languages. However, translationese—characterized by overly literal and unnatural translations—remains a persistent challenge in LLM-based translation systems. 
Despite their pre-training on vast corpora of \textit{natural} utterances, LLMs exhibit translationese errors and generate unexpected \textit{unnatural} translations, stemming from biases introduced during supervised fine-tuning (SFT). 
In this work, we systematically evaluate the prevalence of translationese in LLM-generated translations and investigate its roots during supervised training. 
We introduce methods to mitigate these biases, including polishing golden references and filtering unnatural training instances. 
Empirical evaluations demonstrate that these approaches significantly reduce translationese while improving translation naturalness, validated by human evaluations and automatic metrics. 
Our findings highlight the need for training-aware adjustments to optimize LLM translation outputs, paving the way for more fluent and target-language-consistent translations.
We release the data and code at \href{https://github.com/yafuly/LLM_Translationese}{https://github.com/yafuly/LLM\_Translationese}.

\end{abstract}
\begin{CJK*}{UTF8}{gbsn}
\section{Introduction}
Neural machine translation (NMT) has become the dominant method in machine translation (MT) research~\cite{Attentionis, Edunov:emnlp18, DBLP:journals/corr/abs-1803-05567}.
Recently, advancements in large language models have further expanded the capabilities of NMT, demonstrating notable robustness and generalization across diverse text lengths, structures, and languages~\cite{hendy2023,jiao2023chatgptgoodtranslatoryes,kocmi-federmann-2023-large}.
These works show that LLMs obtain competitive performance on benchmark datasets (e.g., WMT) under automatic metrics, demonstrating strong translation adequacy. 
However, their translation style has been relatively less addressed. 
For example, limited research has been devoted to analyzing and improving the naturalness of translations~\cite{do_literal,iterative_refine}.

\begin{table}[t!]
\centering
\small
\renewcommand{\arraystretch}{1.15}
\begin{CJK*}{UTF8}{gbsn}

\begin{tabular}{p{1cm}p{5.5cm}}
\toprule
\multicolumn{2}{c}{\textbf{Sentence-level Translationese}} \\
\midrule
\textbf{Source} & Few-shot LLMs still lag behind vanilla fine-tuned models \textcolor{blue}{in the task}. \\
\textbf{LLM} & 少样本LLMs仍然落后于原始细化训练模型\textcolor{red}{在任务中}。 (PPL: 151.5) \\
 \textbf{Refine} & \textcolor{red}{在任务中}，少样本LLMs仍然落后于原始细化训练模型。 (PPL: 128.8) \\
\hline
\textbf{Source} & \textcolor{blue}{Bei starker Hitze} ließ diese Festigkeit zwar etwas nach.\\
\textbf{LLM} &  However, \textcolor{red}{at high temperatures} this hardness did diminish somewhat. (PPL: 160.1) \\
\textbf{Refine}  &  However, this hardness did diminish somewhat \textcolor{red}{at high temperatures}.  (PPL: 96.6)\\
\midrule

\multicolumn{2}{c}{\textbf{Phrase-level Translationese}} \\
\midrule
\textbf{Source} & cats \textcolor{blue}  {suffer night blindness} \\
\textbf{LLM} & 猫将\textcolor{red}{遭受夜盲症} (PPL: 335.3) \\
 \textbf{Refine} & 猫会\textcolor{red}{患上夜盲症} (PPL: 154.1) \\
 \hline
\textbf{Source} & \textcolor{blue}{mehr Lebensqualität zu gewinnen}\\
\textbf{LLM} &  \textcolor{red}{gain more quality of life}  (PPL: 620.5)\\
\textbf{Refine}  &  \textcolor{red}{improve the quality of life} (PPL: 27.6)\\
\bottomrule
\end{tabular}
\caption{
\label{tab:intro}
Examples of Sentence-level and Phrase-level Translationese (English-Chinese and German-English translation). 
Source: source text; 
LLM: translations of LLMs;
Refine: translations with translationese refined.
Each case includes an LLM-generated translation alongside a refined version, with perplexity (PPL) values provided at the end. \textcolor{blue}{Blue} text highlights the source segments, while \textcolor{red}{red} text identifies segments in the LLM translation where translationese occurs and is subsequently refined.
}
\end{CJK*}
\end{table}

Existing work shows that machine translation systems can produce less natural translations, a phenomenon known as "translationese"~\cite{2018_translationese_in_nmt,trans_help, dutta2022}.
Translationese occurs when source-language segments are translated too \textit{literally} at either the phrase or sentence level, resulting in deviations from typical target language patterns that sound unnatural to native speakers~\cite{Gellerstam1986, Nida1982}. 
While considerable research has addressed and mitigated translationese in traditional NMT systems~\cite{2018_translationese_in_nmt, Translationese_as_multi}, there has been limited work on whether translationese exists in LLM-based translation systems.

The primary distinction of large translation models lies in the extensive prior knowledge acquired during the pre-training phase, where they learn from a vast corpus of native utterances.
Consequently, LLMs should be less susceptible to translationese patterns and capable of producing natural translations due to their strong language modeling bias.
However, as illustrated in Table~\ref{tab:intro}, LLMs still produce "unexpected" \textit{unnatural} translations despite their exposure to abundant \textit{natural} language data. 
For instance, when translating ``suffer night blindness'' into Chinese, the model generates ``遭受'' as the translation of the word ``suffer'', which is a literal translation but is not typically used for expressing something being afflicted with a disease.

We conduct a systematic evaluation to investigate the translationese patterns exhibited by LLMs and examine the underlying causes of these unexpected unnatural translations, engaging expert translators to meticulously analyze translationese in LLMs.
Initially, we collect documents from diverse writing domains
and use both translation-specialized (e.g., ALMA~\cite{alma2}) and general LLMs (e.g., GPT4~\cite{gpt4}) for generating translations.
For each translated document, expert translators identify specific spans exhibiting pre-defined translationese error types.
We then compute the proportion of these spans, termed the Translationese Span Ratio (TSR), and average these ratios across annotators to provide a quantitative measure of translationese prevalence.

Results indicate that all LLMs exhibit significant translationese errors in both English-Chinese and German-English translations. 
Notably, even advanced models like GPT-4 demonstrate over 40\% of their translations as exhibiting substantial translationese patterns.
Interestingly, when LLMs are asked to refine their own translations, they produce more natural outputs with markedly lower TSRs. 
For example, in Table~\ref{tab:intro}, after refining the translation, ``suffer'' becomes ``患上'' .
This suggests that LLMs own prior knowledge and potential for generating natural translations, but may be biased during supervised training (i.e., supervised fine-tuning, SFT) for the ``translation'' task, placing excessive emphasis on literal semantic mapping at the expense of fluent language generation.

We validate LLMs' potential of generating natural translations by demonstrating a positive correlation between their predicted perplexities and human evaluation: higher perplexities are often associated with increased TSRs. 
As shown in Table~\ref{tab:intro}, the perplexities of direct LLM translations are higher than those of the refined ones. 
This finding not only verifies our hypothesis above to some extent but also provides an automatic metric for detecting translationese.
To further verify biases introduced during supervised fine-tuning (SFT), we engage expert translators to analyze translationese in sampled training instances from widely used SFT datasets.
Our findings reveal that over 34\% of these training instances exhibit translationese patterns, indicating that LLMs may be biased towards producing unnatural translations during SFT.

We propose two mitigation strategies to address translationese. First, LLMs’ natural potential is leveraged to refine golden training references, reducing translationese patterns. Empirical evaluations on Llama-3.1-8B and Qwen-2.5-7B show that refining training instances improves translation naturalness significantly, as confirmed by both automatic and human evaluations. Second, pre-trained LLMs are used to filter unnatural translations from supervised fine-tuning (SFT) data, which also enhances translation naturalness. 
Extensive experiments across additional languages further demonstrate the generalizability of our method. 
To our knowledge, this is the first systematic study addressing translationese in LLMs. 
We will release our resources after the anonymous period.


\end{CJK*}

\section{Related Work}

\paragraph{Translationese in Machine Translation.}
Translationese refers to the phenomenon in which translated texts display linguistic characteristics that diverge from the typical patterns of the target language, resulting in overly literal expressions that sound unnatural to native speakers~\citep{Gellerstam1986, Nida1982}. 
A line of work has explored translationese and proposed dedicated mitigation strategies.
\citet{trans_help} analyze the translationese by measuring various linguistic features, while \citet{level_expert} find that texts with translationese elicit higher perplexities.
Several studies have identified data quality issues as a contributing factor to translationese. 
Researchers~\citep{post-editese, zhang-toral-2019, ni_etal_2022_translationese_and_perform, wang_etal_2023} study the impact of translationese on model performance, whereas another line of work~\citep{Translationese_as_multi, translating_away, low_resource, translationese_data_augmentation} relies on translationese to enhance data quality or achieve data augmentation.
\citet{dutta2022} and \citet{wein2024} propose to address the translationese issue using specialized algorithms, while \citet{translationese_prompt_engi} focus on prompt-engineering to mitigate this issue.
Unlike their work, we focus on the unexpected translationese in the context of powerful LLMs. 




\paragraph{Large Language Model for Translation.}


Recent studies demonstrate the strong translation capabilities of LLMs like GPT-3.5 and GPT-4, particularly with in-context few-shot learning~\cite{jiao2023chatgptgoodtranslatoryes, hendy2023, kocmi-etal-2023-acl, alam1, zhu-etal-2024}. 
A line of work enhances translation performance through prompt engineering, such as dictionary-based approach \citep{2023dictionary_based}, knowledge extraction by self-prompting \citep{2023_knowledge_extraction} or self-evaluation and refinement \citep{feng_2024_self_correction, ki_carpuat_2024_acl, iterative_refine}. 
From a training perspective, researchers~\citet{NIPS_2022_instruction_tuning}, \citet{jiao_2023-ins_tuning}, \citet{Zeng2023TIMTL} and~\citet{2024_apple_ins_tuning} propose instruction tuning methods to enhance model alignment with human feedback by comparing multiple translations.
~\citet{lexmatch} propose a dictionary-based data curation method for efficient SFT.
\citet{alma2} identify data quality issues in SFT as a potential contributor to suboptimal translation performance, further corroborated by findings from \citet{2024_preference_alignment_best_option}.

LLMs have excelled in producing fluent and adequate translations, effectively addressing faithfulness and accuracy. However, achieving stylistically natural translations remains a significant challenge.
While \citet{do_literal} report a reduction in overly literal translations from LLMs, unnatural expressions still pose a significant challenge~\cite{iterative_refine}. 
In this work, we systematically analyze the origins of LLM translationese and propose training-aware mitigation methods.



\section{Translationese in LLM Translation}
To gain a systematic and quantitative assessment of translationese errors in LLM translation, we perform fine-grained human annotation on the outputs generated by these models based on source documents from typical writing tasks.

\begin{figure*}[t!]
    \centering
    \begin{subfigure}[b]{\textwidth}  
        \centering
        \includegraphics[width=\textwidth]{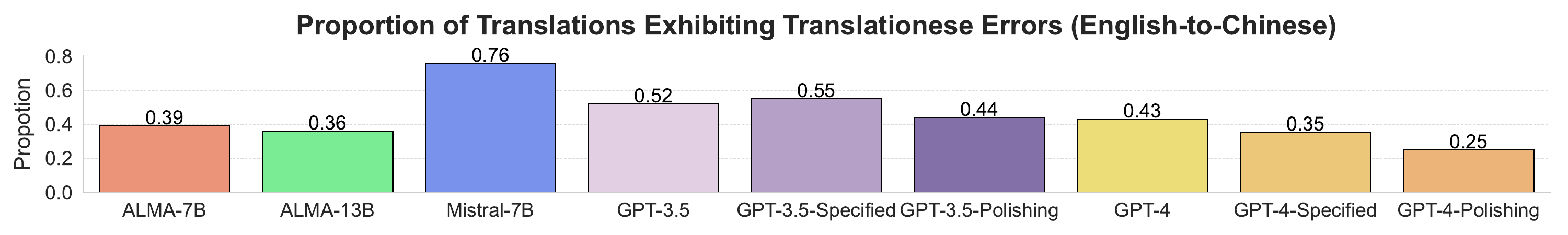}  
        \label{fig:tsr1}
    \end{subfigure}
    
    \vspace{-1em}  

    \begin{subfigure}[b]{\textwidth}  
        \centering
        \includegraphics[width=\textwidth]{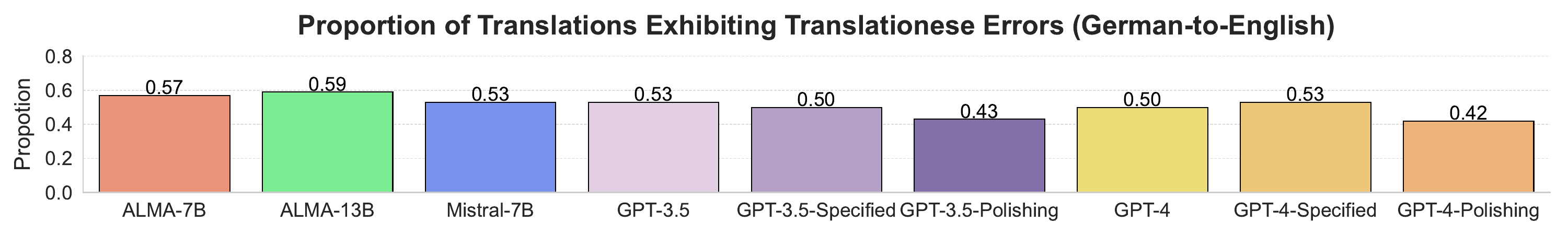}  
        \label{fig:tsr2}
    \end{subfigure}

    \caption{Proportions of translations exhibiting translationese errors. All LLMs adopt direct translation prompts, with the exception of GPT-3.5 and GPT-4, which incorporate supplementary prompts to facilitate more natural translations. Both ``Specified'' and ``Polishing'' prompts have identical requirements; however, the `Polishing' prompt specifically instructs LLMs to refine their generated translations.}
    \label{fig:tsr}
\end{figure*}

\subsection{Data Collection}
We examine four writing domains: news articles, scientific writings, Wikipedia entries, and social media comments. 
We consider English-Chinese (En-Zh) and German-English (De-En) translations.
For the English source segments, we web-crawled 50 document-level samples from each of the following sources: CNN News\footnote{https://www.cnn.com/}, Arxiv\footnote{https://arxiv.org/}, Wikipedia\footnote{https://www.wikipedia.org/}, and Quora forums\footnote{https://www.quora.com/}. 
This process results in 200 English source documents.
For the German source segments, we obtained 100 document-level samples consisting of news articles from Focus\footnote{https://www.focus.de/} and comments from Quora forums.

We employ both commercial LLMs such as GPT-3.5-Turbo and GPT-4-Turbo~\cite{gpt4} as well as open-source alternatives including ALMA-7B-R, ALMA-13B-R~\cite{alam1,alma2}, and Mistral-7B-Instruct-v0.3~\cite{mistral}. 
ALMA models are specialized translation models while the other models are general chat models\footnote{Model selection is based on our empirical studies of document-level translation ability.}.
All the models employ a straightforward translation prompt, with the exception of GPT models, which use two variants to mitigate translationese errors: the \textbf{specified} prompt and the \textbf{polishing} prompt.
While both prompts have the same requirements focused on the target language style, the polishing prompt specifically requires refinement of an existing translation, which is a two-step process: first performing direct translation followed by polishing, as detailed in Appendix~\ref{app:prompt}.

In this way, each document is translated using nine models: ALMA-7B, ALMA-13B, Mistral-7B, GPT-3.5, GPT-3.5-Specified, GPT-3.5-Polishing, GPT-4, GPT-4-Specified, and GPT-4-Polishing, where ``Specified'' and ``Polishing'' refer to using the respective prompts. This process yields a total of 1,800 document-level English-Chinese translations and 900 German-English translations for human annotation, as summarized in Appendix~\ref{app:statistics}.

\subsection{Translationese Span Annotation}
\label{sec:anno}
Using Label Studio~\cite{label_studio}, we develop a specialized annotation platform to help expert translators identify text spans with translationese errors.
Inspired by Unbabel's annotation guidelines,
we categorize translationese errors into two primary types: \textbf{unnatural sentence flow} and \textbf{unnatural phrase flow}, corresponding to sentence-level and phrase-level translationese.
Unnatural sentence flow occurs when source language structures are translated directly without adequate adaptation to the target language, 
whereas unnatural phrase flow pertains to overly literal translations of source phrases.
Recognizing that traditional translation errors (e.g., omissions and mistranslations) can also occur in LLM outputs, we include these types of errors in our annotation guidelines and platform.
Based on the aforementioned translation error taxonomy, we request three expert translators to identify and annotate segments containing translation errors, specifically focusing on two types of translationese errors.
The annotators, all of whom hold advanced degrees in linguistics or translation studies and possess extensive experience in professional translation, ensure a high level of accuracy and consistency in identifying nuanced translation errors.
Detailed annotation guideline and platform demonstration can be found in Appendix~\ref{app:guide}.

\subsection{Human Evaluation Results}
\label{sec:anno_res}

We gather human annotation results and calculate the length ratio of spans exhibiting translationese errors (i.e., unnatural sentence and phrase flow) for each document, termed the \textbf{translationese span ratio} (TSR). 
For example, a TSR of 0.2 signifies that 20\% of the documents exhibit translationese. 
The TSRs from three translators are averaged for each document, and then aggregated across all translations for each model.
To complete the fine-grained TSR metric, we evaluate the \textbf{proportion} of documents with significant translationese errors (significant errors are defined as a TSR greater than 0.2). 
These documents (TSR$>$0.2) represent translations that are notably unnatural from a native speaker's perspective.
We demonstrate this document-level analysis in Figure~\ref{fig:tsr}.
Direct TSR scores are presented in Appendix~\ref{app:tsr}. 

\paragraph{Overall Results.}

As shown in Figure~\ref{fig:tsr}, all large language models display significant translationese patterns in both English-Chinese and German-English translations, with an average of 45.0\% and 51.1\% of document-level translations displaying translationese for English-Chinese and German-English translations, respectively.
We first examine model translations under the ``direct'' translation prompt setting.
For English-Chinese translation, larger models generate more natural translations (GPT4 v.s. GPT3.5 and ALMA-13B v.s. ALMA-7B),
and specialized translation models (ALMA) generate fewer translationese errors compared to general chat models like Mistral-7B, GPT-3.5, and GPT-4. 
For instance, ALMA-13B produces 36.0\% of documents with translationese, whereas the lowest-performing model, Mistral-7B, exhibits a rate of 76.0\%. 
For German-English translation, all models demonstrate minimal variati
on.
This discrepancy may stem from the fact that most LLMs are pre-trained on an unbalanced corpus dominated by English, with significantly varying proportions of other languages.
Regarding types of translationese errors, unnatural sentence flow errors occur more frequently than unnatural phrase flow errors; averaged error annotation counts are 3549.0 versus 1690.0 for English-Chinese translations and 1655.0 versus 311.7 for German-English translations.
Examples of translationese cases can be found in Appendix~\ref{app:case}.

\paragraph{Prompting LLMs for Reducing Translationese.}
We explore the effects of the two alternative prompts: ``specified'' and ``polishing'' prompt.
Interestingly, incorporating specific requirements (i.e., ``specified'') in prompts that intend to enhance naturalness does not consistently reduce the rate of translationese errors; in some cases, it may even worsen the translation quality. 
For instance, under specified prompts, GPT-4 exhibits an increase in translationese errors, with the proportion rising from 0.50 to 0.53.
Conversely, refining translations generated by the LLM itself (``polishing'') effectively and steadily reduces translationese errors.
In particular, GPT-4 decreases the proportion of translationese from 43\% to 25\% through self-polishing its own translations.
This indicates that it is not style-constrained prompts that promote natural generation but rather the task formats themselves, namely ``translate'' and ``polishing''.
In other words, \textit{while LLMs pre-trained on extensive native utterances can generate more natural translations, this potential is not realized within a "translation" prompt.} 
The subsequent sections will explore the supervised training phase, where LLMs are instructed to perform various generation tasks, to investigate the origins of ``unexpected'' \textit{unnatural translations} they generate despite their exposure to massive amounts of \textit{natural} language during pre-training.


\section{Tracing Translationese in Supervised Training Data}
\begin{figure}[!t]
\centering
\setlength{\belowcaptionskip}{-0.2cm}   
\includegraphics[width=0.9\linewidth]{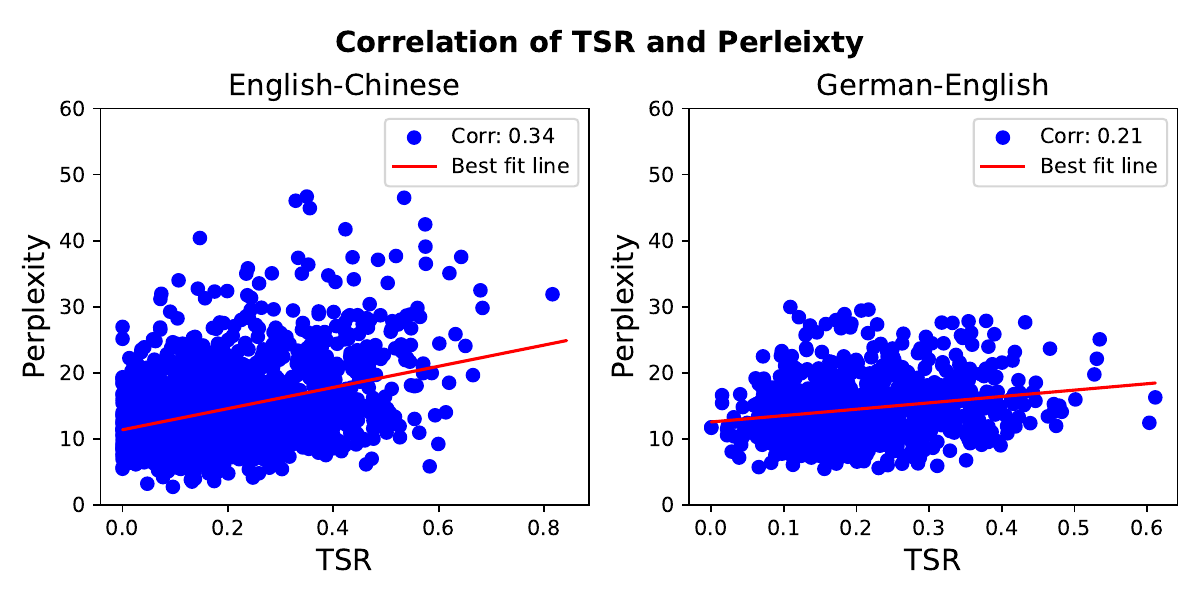}
\caption{Correlation between the human-annotated translation span ratio (TSR) and LLM-generated perplexity.
}
\label{fig:corr_ppl_tsr}
\end{figure}

To investigate the origins of unnatural translations produced by LLMs, 
we first analyze the inherent preference of LLMs for natural generations and subsequently examine potential biases introduced during supervised training.
We contend that LLMs trained on extensive corpora have the potential to distinguish unnatural generations, offering a reliable sign of generation naturalness. 
Previous studies~\cite{trans_help,level_expert,translating_away,low_resource} use target language model perplexity as a metric for translationese, where higher perplexity indicates less natural generation. 
However, these studies rely on language models trained on limited target-language corpora. 
In this work, we employ Llama-3.1-8B~\cite{llama3}, a large language model pre-trained on vast multilingual data that exhibits exceptional multilingual capabilities, to assess generation naturalness.
Specifically, we calculate the perplexity of each translation, excluding the source text context, using Llama-3.1-8B and analyze its correlation with the human-annotated translation span ratio (TSR).
As illustrated in Figure~\ref{fig:corr_ppl_tsr}, despite being measured at different granularities (document-level versus span-level), these two metrics exhibit a positive correlation, particularly evident in English-Chinese translations, where higher perplexity corresponds to an increased ratio of spans identified as translationese errors.

\begin{figure}[!t]
\centering
\setlength{\belowcaptionskip}{-0.2cm}   
\includegraphics[width=0.9\linewidth]{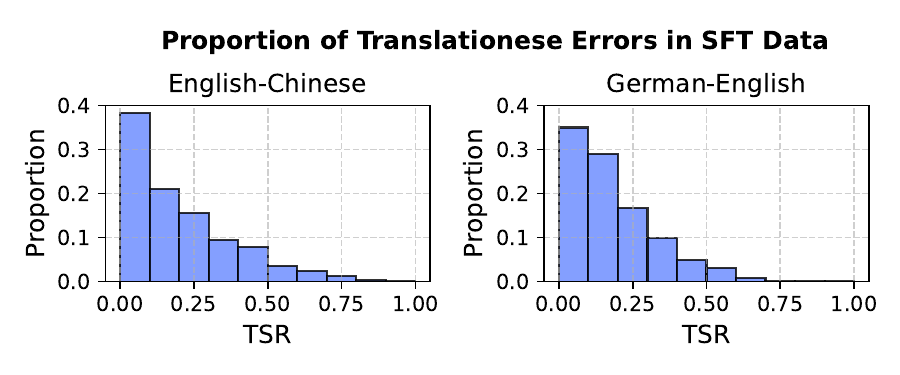}
\caption{
Proportions of supervised training instances exhibiting different levels of translationese errors (TSR).
}
\label{fig:sft_tsr}
\end{figure}

We hypothesize that biased data in supervised training significantly contributes to translationese patterns, even though pre-trained LLMs favor natural sequences.
As suggested by previous work~\cite{alam1,alma2}, supervised training data for LLM translation systems consists of test and validation data from existing benchmark datasets (e.g., WMT and Flores~\cite{no_lang}). 
However, these test datasets still exhibit translationese errors~\cite{zhang-toral-2019}, potentially introducing biases during supervised training.
To quantify these biases, we sample 500 instances of English-Chinese and German-English translations from the ALMA training set~\cite{alam1,alma2}, asking the three expert translators to annotate the translationese spans for each instance (Details in Appendix~\ref{app:guide_sent}).
Translation span ratios from the 3 annotators are computed and averaged, with results shown in Figure~\ref{fig:sft_tsr}.
A notable percentage of sentences contains over 20\% spans identified as translationese: 40.4\% for English-Chinese and 34.2\% for German-English instances.
The majority of errors stem from overly literal translation patterns, causing unnatural sentence- or phrase-level flows.
This suggests that during supervised training, the LLM may develop a bias towards interpreting the "translation" task as a direct transformation from source to target, overemphasizing faithfulness at the expense of naturalness.

\section{Mitigating Translationese from Supervised Training}

In this section, we validate our hypothesis by addressing translationese biases in SFT and empirically evaluating translation naturalness.

\subsection{Training Settings}
\label{sec:sft}
We primarily adopt the training configurations from ALMA~\cite{alam1} to develop LLMs for English-Chinese and German-English translation.
For parallel training data, we extract instances for both translation directions (En-Zh and De-En) from the ALMA training set (WMT'17 to WMT'21 and Flores-200~\cite{no_lang}), resulting in a total of 31,621 parallel training instances.
To construct the development set, we randomly select 10\% of the training data.
For evaluation, we assess models using our collected \textbf{document-level} datasets as well as \textbf{sentence-level} test sets from WMT'22.
We use Llama-3.1-8B and Qwen-2.5-7B~\cite{qwen} as base models due to their superior multilingual capabilities. 
Training details are presented in Appendix~\ref{app:train}.
\begin{table*}[t!]
\centering
\small
\renewcommand{\arraystretch}{1.2}
\setlength{\tabcolsep}{4.5pt}
\begin{tabular}{c|ccc|ccc|ccc|ccc}
\toprule
\multirow{3}{*}{\textbf{Training}} & \multicolumn{6}{c|}{\textbf{Document-level Translation}} & \multicolumn{6}{c}{\textbf{Sentence-level Translation}} \\
 & \multicolumn{3}{c|}{\textbf{En-Zh}} & \multicolumn{3}{c|}{\textbf{De-En}} & \multicolumn{3}{c|}{\textbf{En-Zh}} & \multicolumn{3}{c}{\textbf{De-En}} \\ 
 & Lex.$\uparrow$ & Len.$\uparrow$ & PPL$\downarrow$ &Lex.$\uparrow$ & Len.$\uparrow$ & PPL$\downarrow$ &Lex.$\uparrow$ & Len.$\uparrow$ & PPL$\downarrow$ &Lex.$\uparrow$ & Len.$\uparrow$ & PPL$\downarrow$ \\
 \midrule
 & \multicolumn{12}{c}{Llama-3.1-8B}  \\
 \midrule
SFT & 
\colorbox{lowblue}{0.509} & \colorbox{lowblue}{0.639} & 13.8 & \colorbox{lowblue}{0.421} & 0.079 &  \colorbox{lowblue}{15.0} & \colorbox{lowblue}{0.500} & \colorbox{lowblue}{0.377} &  103.3  & 
\colorbox{lowblue}{0.415} & \colorbox{lowblue}{0.150} & 84.2 \\

SFT-KD  & 
\colorbox{lowblue}{0.509} & 0.648 & \colorbox{lowblue}{14.3} & 
0.424  & \colorbox{lowblue}{0.078} &  14.4 & 
0.503 & 0.406 &  \colorbox{lowblue}{104.9}  & 
\colorbox{lowblue}{0.415}  & 0.153 & \colorbox{lowblue}{88.1} \\

SFT-Polished & 
\colorbox{lowred}{0.522} & \colorbox{lowred}{0.717} &  \colorbox{lowred}{11.9} & 
\colorbox{lowred}{0.438} & \colorbox{lowred}{0.080} &  \colorbox{lowred}{13.8} & 
\colorbox{lowred}{0.514} &  \colorbox{lowred}{0.466} & \colorbox{lowred}{90.0} & 
\colorbox{lowred}{0.419} & \colorbox{lowred}{0.165} &  \colorbox{lowred}{72.7} \\
\midrule

&  \multicolumn{12}{c}{Qwen-2.5-7B}  \\
 \midrule
SFT &
\colorbox{lowblue}{0.511} & \colorbox{lowblue}{0.600} & 13.8 & 
\colorbox{lowblue}{0.418} & \colorbox{lowred}{0.077} & \colorbox{lowblue}{14.8} & 
0.508 & 0.279 & 101.6& 
\colorbox{lowblue}{0.409} & 0.136 & \colorbox{lowblue}{88.8} \\

SFT-KD & 
0.513 & 0.651 & \colorbox{lowblue}{13.9} & 
0.424 & \colorbox{lowblue}{0.068} & 14.7 &  
\colorbox{lowblue}{0.505} & \colorbox{lowblue}{0.272}  & \colorbox{lowblue}{104.2} & 
0.415 & \colorbox{lowblue}{0.129} & 88.4\\

SFT-Polished & 
\colorbox{lowred}{0.523} & \colorbox{lowred}{0.687} & \colorbox{lowred}{12.1} & 
\colorbox{lowred}{0.436} & 0.073 & \colorbox{lowred}{14.3} & 
\colorbox{lowred}{0.518} & \colorbox{lowred}{0.317} & \colorbox{lowred}{87.3} & 
\colorbox{lowred}{0.419} & \colorbox{lowred}{0.139} & \colorbox{lowred}{71.1}\\
%
\bottomrule
\end{tabular}
\caption{Automatic evaluation of translation naturalness at both sentence and document levels across different training methods, where a \colorbox{lowred}{red} background indicates the best performance and a \colorbox{lowblue}{blue} one signifies the worst.}
\label{tab:main}
\end{table*}
\begin{table}[t]
\centering
\small
\begin{tabular}{cccc}
\toprule
\textbf{Direction} & SFT & SFT-KD & SFT-Polished \\ 
 \midrule
\textbf{En-Zh} & 2.3 & 2.2 & \textbf{1.4} \\
\textbf{De-En} & 2.3 & 2.0 & \textbf{1.7} \\
\bottomrule
\end{tabular}
\caption{Average ranks for various SFT methods. Lower values indicate better performance.}
\label{tab:votes}
\end{table}

\begin{table}[t!]
\centering
\small
\begin{tabular}{ccccc}
\toprule
\multirow{2}{*}{\textbf{Training}} &\multicolumn{2}{c}{\textbf{Llama-3.1-8B}}  & \multicolumn{2}{c}{\textbf{Qwen-2.5-7B}}\\
& \textbf{En-Zh} & \textbf{De-En} & \textbf{En-Zh} & \textbf{De-En} \\ 
 \midrule
SFT  & 80.0  & 80.5 & 73.8 & 74.0\\
SFT-KD & 81.5  & \textbf{81.2} & \textbf{74.7} & 75.3\\
SFT-Polished & \textbf{81.8} & 81.0  & 74.2 & \textbf{75.6}\\
\bottomrule
\end{tabular}
\caption{Translation quality evaluation (COMET-QE).}
\label{tab:quality}
\end{table}

\begin{table}[t]
\begin{CJK*}{UTF8}{gbsn}
\centering
\small
\begin{tabular}{p{1cm}p{5.5cm}}
\toprule
 \multicolumn{2}{c}{English-to-Chinese}\\
 \midrule
 \textbf{Source} & I've looked into it and I can see that your area is currently having a high volumes of order that is \textcolor{blue}{why they were assigning a rider for your order}. \\
 \midrule
\textbf{SFT} & 我已经调查过了，你的地区订单量非常大，\textcolor{red}{才会把骑手分配给你的订单}。  \\
\textbf{SFT-KD} & 我已经调查过了，你的地区当前订单量很大，\textcolor{red}{这就是为什么他们会为你的订单安排骑手的原因}。 \\
\textbf{SFT-Polished} & 我已经调查了情况，你的地区当前订单量很大，\textcolor{red}{因此才有骑手为你配送订单}。 \\
\midrule
 \multicolumn{2}{c}{German-to-English}\\
 \midrule
 \textbf{Source} & Dank der VET-Technologie (Virtual Element Technology), die auf der Analyse der Bauteile und Amp-Schaltungen beruht, \textcolor{blue}{warten die VTX-Verstärker mit einer Authentizität auf, welche absolut beeindruckend ist}. \\
 \midrule
\textbf{SFT} & Thanks to VET technology (Virtual Element Technology) based on the analysis of the parts and amp circuits, \textcolor{red}{the VTX amplifiers wait with an authenticity that is absolutely staggering.}\\
\textbf{SFT-KD} & Thanks to VET technology (Virtual Element Technology), which is based on the analysis of components and amp circuits, \textcolor{red}{the VTX amplifiers offer an authenticity that is absolutely impressive.} \\
\textbf{SFT-Polished} & Thanks to VET technology (Virtual Element Technology), which is based on the analysis of components and amplifier circuits, \textcolor{red}{the VTX amplifiers deliver a level of authenticity that is truly astounding.}\\
\bottomrule
\end{tabular}
\caption{Case study of different model translations.}
\label{tab:case_study}
\end{CJK*}
\end{table}

\subsection{Evaluation Metrics}
We use both automatic and human evaluation metrics to assess the translation naturalness.

\paragraph{Automatic Evaluation.}
As discussed, \textbf{perplexity} (PPL) is an effective indicator of generation naturalness~\cite{translating_away,low_resource}.
Following previous work~\cite{trans_help,zhang-toral-2019,translating_away,Translationese_as_multi}, we consider two additional metrics: \textbf{lexical density} (Lex.) and \textbf{length variance} (Len.). 
Lexical density is defined as the ratio of content words to total words, as translationese typically exhibits lower lexical complexity and a reduced proportion of content words (adverbs, adjectives, nouns, and verbs)~\cite{scarpa2006corpus}.
We use Stanza~\cite{stanza} to extract part-of-speech tags and content words accordingly.
Both machine translation (MT) systems and human translators typically refrain from restructuring the source sentence, adhering instead to prevalent sentence structures in the source language. Consequently, this practice yields translations that closely match the length of the original sentences.
For each source-target pair $(x,y)$, the length variety is calculated as: $\frac{||x|-|y||}{|x|}$.
For translation quality estimation, we utilize \texttt{Unbabel/wmt22-cometkiwi-da} to compute and report COMET-QE scores~\cite{comet}. 
We choose reference-free scores to avoid possible translationese biases in the reference translations from the test set.

\paragraph{Human Evaluation.}
\label{sec:human_rank}
We ask the three expert translators to rank translations generated by different models in accordance with the annotation guidelines outlined in Section~\ref{sec:anno}. Unlike previous tasks, their focus is solely on ranking translations rather than identifying fine-grained spans (Details in Appendix~\ref{app:human_rank}).

\subsection{Improving Naturalness of Training Data}
As suggested in Section~\ref{sec:anno_res}, using LLMs to polish existing translations can enhance translation naturalness. 
To mitigate translationese bias in SFT data, we use the polishing prompt to let GPT-4 refine the golden references (Appendix~\ref{app:prompt}). 
Subsequently, we fine-tune LLMs with these polished translations, referred to as ``\textbf{SFT-Polished}''.
Additionally, to ablate knowledge distillation from GPT-4, we use GPT-4 to generate direct translations of the source training instances, termed ``\textbf{SFT-KD}''.
Table~\ref{tab:main} compares translation naturalness between the baseline ``SFT'' method and other approaches.

As shown in the Table, addressing translationese bias in SFT data effectively mitigates model translationese for both base LLMs, with SFT-Polished yielding consistent improvements across all automatic metrics, i.e., higher lexical densities, increased length variability, and reduced perplexities. 
Specifically, the perplexities of translations from SFT-Polished are significantly lower than those from SFT and SFT-KD ($p<0.01$), with average reductions of 7.8 for English-Chinese and 7.7 for German-English translations.
In contrast, direct knowledge distillation from GPT-4 fails to enhance translation naturalness and may even degrade it in certain cases.
This finding suggests that using LLMs such as GPT-4 to directly translate training data can not rectify existing translationese bias, as these LLMs may already be influenced by biases introduced during supervised training for translation tasks.
Nevertheless, LLMs can improve naturalness through alternative task formats such as polishing.

As shown in Table~\ref{tab:votes}, human evaluations of translations from models fine-tuned on Llama-3.1-8B corroborate the automatic assessments: SFT-Polished achieves the highest rankings and demonstrates strong inter-annotator agreement in both directions (details regarding inter-annotator agreement are provided in Appendix~\ref{app:human_rank}).
Translation quality estimation on the WMT test sets, as shown in Table~\ref{tab:quality}, indicates that both SFT-KD and SFT-Polished significantly enhance translation quality ($p<0.01$). 
Table~\ref{tab:case_study} highlights the improvements achieved by SFT-Polished, such as transforming overly literal German-to-English translations like “wait with an authenticity ” into the more stylistically natural “deliver a level of authenticity” (see Appendix~\ref{app:case2} for additional examples).

\begin{figure}[!t]
\centering
\includegraphics[width=0.99\linewidth]{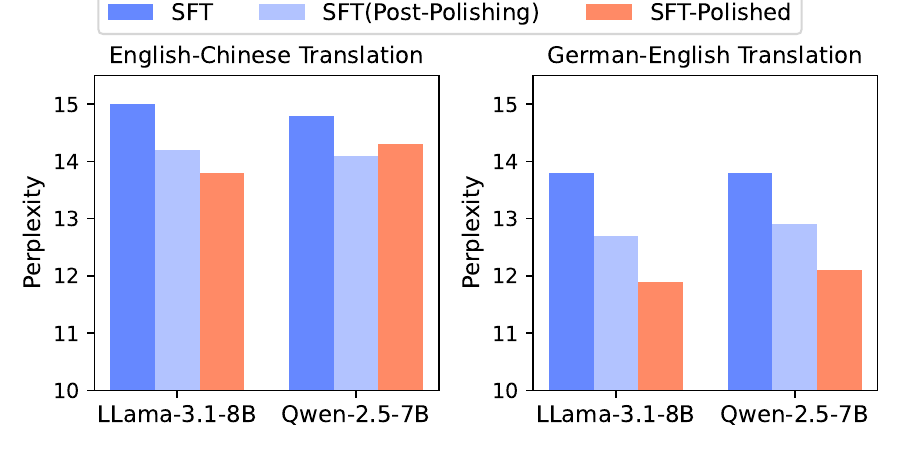}
\caption{
Comparison of naturalness between inference-time (Post-Polishing) and training-time polishing (Polished).
}
\label{fig:compare_post}
\end{figure}

Additionally, we compare SFT-Polished models, which are trained on polished data, with SFT-Post-Polishing models that employ GPT-4 to refine translations produced by SFT models.
As shown in Figure~\ref{fig:compare_post}, incorporating polishing during both training and inference improves translation naturalness, as indicated by reduced perplexities.
Nevertheless, training on polished training instances results in more substantial improvements in translation naturalness, further supporting our hypothesis that translationese is predominantly shaped during supervised training.

\begin{figure}[t]
\centering
\includegraphics[width=0.99\linewidth]{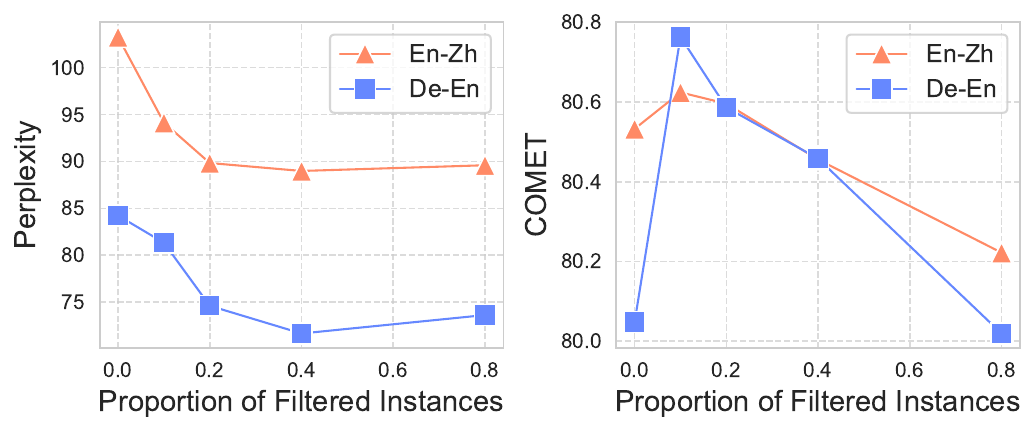}
\caption{Translation naturalness and quality w.r.t. filtered training samples.
}
\label{fig:sft_filter}
\end{figure}

\subsection{Filtering Unnatural Training Instances}
An alternative approach to mitigate translationese bias involves filtering out unnatural training references before supervised training. 
We take perplexity as a measure of naturalness, allowing us to rank training instances and exclude the least natural subset.
Experiments are conducted using Llama-3.1-8B.
The results are illustrated in Figure~\ref{fig:sft_filter}, which displays the relationship between translation naturalness and quality on sentence-level WMT test sets relative to the proportion of filtered training instances.
As shown in Figure~\ref{fig:sft_filter}, filtering up to 40\% of the least natural references consistently enhances translation naturalness.
Moreover, moderate filtering also improves translation quality. 
Specifically, a filtering proportion of 20\% yields improvements in both metrics.
However, excessive filtering adversely affects both naturalness and quality.

\subsection{Generalization to More Languages}

\begin{table}[t]
\setlength{\belowcaptionskip}{-0.2cm}
\centering
\small
\begin{tabular}{ccccc}
\toprule
\textbf{Training} & \textbf{En-Is} & \textbf{En-Cs} & \textbf{En-De}  & \textbf{En-Ru} \\ 
 \midrule
 \multicolumn{5}{c}{Perplexity$\downarrow$} \\
 \midrule
SFT   & 27.0& 59.9  & 56.5 & 42.8\\
SFT-Polished & \textbf{24.9} & \textbf{50.9} & \textbf{44.0}   & \textbf{35.9}\\
 \midrule
 \multicolumn{5}{c}{COMET-QE$\uparrow$} \\
 \midrule
 SFT   & 80.6	& 81.3	& 63.0	 & 81.0 \\
SFT-Polished & \textbf{84.1} & \textbf{83.1} & \textbf{65.7} & \textbf{82.4} \\
\bottomrule
\end{tabular}
\caption{Generation naturalness (perplexity) and quality (COMET-QE) of translations from English to four additional languages.}
\label{tab:gen_lang}
\end{table}

We extend our hypothesis to additional languages and evaluate the effectiveness of SFT-Polished.
Specifically, we focus on translating from English to two high-resource languages: German (De) and Russian (Ru), as well as two moderate-resource languages: Czech (Cs) and Icelandic (Is). We use the same training and test sets from ALMA~\cite{alam1}.
To train a multilingual translation model based on Llama-3.1-8B. We combine the additional training data with the original training set in Section~\ref{sec:sft}. The naturalness of translations for these four languages is presented in Table~\ref{tab:gen_lang}. 
SFT-Polished generates translations with an average perplexity decrease of 7.6. 
In particular, the perplexity decreases from 56.5 to 40.0 for English-German translation.
Our results demonstrate that polishing the training data consistently and significantly ($p<0.01$) reduces translationese bias across all four languages, yielding a more natural translation.
In addition, SFT-polished obtains consistently better translation quality compared with the SFT counterparts.

\section{Conclusion}
In this work, we revealed how translationese, a long-standing issue in machine translation, persists even in state-of-the-art LLMs due to biases introduced during supervised training. 
Systematic analysis demonstrated the high prevalence of unnatural translations across multiple models and language pairs, attributed to training data with inherent translationese patterns. 
By leveraging techniques such as refining golden references and filtering unnatural instances, we achieved significant improvements in translation naturalness, confirming the potential of LLMs to align closer to native linguistic patterns. 
These findings underscored the importance of addressing data quality and training methodologies in developing robust and natural translation systems. 
Future research should extend these approaches to a broader range of language pairs and domains.


\section*{Limitations}
While this study provides valuable insights into the issue of translationese in LLM-generated translations, several limitations should be acknowledged. 
First, due to the significant costs in time and resources required for human annotations, the evaluation primarily focuses on English-Chinese and German-English translations, which may limit the generalizability of the findings to other language pairs, especially low-resource or morphologically rich languages. 
Second, despite efforts to include a broad range of LLM translation systems, there are still other models and architectures that warrant further exploration. 
Third, while our findings reveal that SFT introduces significant translationese bias, translationese can also stem from other training phrases, such as pre-training and reinforcement learning, which we leave for future work.
Finally, while human and automatic evaluations are employed, subjective biases in human annotations and the limitations of current automatic metrics could influence the assessment of translation naturalness. Addressing these limitations in future work could enhance the robustness and applicability of the findings.

\section*{Ethic Considerations}
The data utilized in this study is web-crawled from publicly available sources, or obtained from publicly available datasets designed for academic research and contains no sensitive information. These datasets, including sources such as WMT and Flores, are freely accessible for non-commercial use, and their legality for academic purposes has been confirmed by our institution’s legal advisors.

Our data construction involves human annotations to identify translationese patterns (Section~\ref{app:guide} and Section~\ref{app:guide_sent}) and rank LLM translations (Section~\ref{app:human_rank}). 
All annotators are tasked with reviewing translations, ensuring that no personal or sensitive information is included in the process. 
Three expert translators with advanced degrees in Linguistics or related fields are hired for annotation work of both translation directions. 
Before conducting formal annotations, they undergo a training phase that includes annotating 100 samples to ensure consistency and accuracy. Subsequently, they completed the aforementioned formal annotation tasks. Annotators are paid for both their training and formal annotation work at a rate of \$16 per hour, determined based on the average annotation time for the training samples. 
This rate is designed to ensure fair and ethical compensation. Each annotator spends a total of 216 hours on the annotation (for English-Chinese), or 192 hours (for German-English), with compensation of \$3,456 or \$3,072, respectively.

No datasets are created that involve unethical content, and we make every effort to remove any data points that could potentially cause ethical concerns. 
We comply with the terms set by companies offering commercial LLM APIs and extend our gratitude to all collaborators for their invaluable support in utilizing these APIs.
Additionally, our findings and methodologies aim to improve translation quality and do not promote harmful or biased content generation. By adhering to these standards, we ensure that this study was conducted ethically and responsibly.

\bibliography{custom}

\newpage

\appendix

\section{Translation Prompt}

We employ three types of prompts for translations using large language models.
As illustrated in Table~\ref{tab:prompt}, all models utilize the basic translation prompt; however, the well-instructed GPT models (GPT-3.5 and GPT-4) incorporate two additional prompts: the specified prompt and the polishing prompt.
\label{app:prompt}

\begin{table*}[t]
    \small
    \centering
    \begin{tabular}{p{0.2\linewidth}p{0.8\linewidth}}
    \toprule
        \textbf{Translation Prompt} &  Please translate the following \{source\_language\} text to \{target\_language\}.  \\
        &\#\#\# Source text: \{source\_text\} \\
        &\#\#\# Translation:
        \\
    \midrule
        \textbf{Specified Prompt} &  Please translate the following \{source\_language\} text to \{target\_language\}, ensuring that the translation is fluent, accurate, and conforms to typical \{target\_language\} expressions and style.  \\
        &\#\#\# Source text: \{source\_text\} \\
        &\#\#\# Translation: \\
    \midrule
        \textbf{Polishing Prompt} & Please polish the corresponding \{target\_language\} translation of an \{source\_language\} text, ensuring that the translation is fluent, accurate, and conforms to typical \{target\_language\} expressions and style. \\
        &\#\#\# Source text: \{source\_text\} \\
        &\#\#\# Original Translation: \{target\_text\} \\
        &\#\#\# Translation: \\
    \bottomrule
    \end{tabular}
    \caption{Three types of prompts used in large language model translation. The first one is utilized for all models whereas the other two are only used in GPT models.}
    \label{tab:prompt}
\end{table*}

\section{Data Statistics}
\label{app:statistics}
\begin{table}[t!]
    \small
    \centering
    \begin{tabular}{p{0.15\linewidth}p{0.2\linewidth}p{0.15\linewidth}p{0.15\linewidth}}
    \toprule
        Direction & Domains & Avg. Tokens & \#. Docs. \\
    \midrule
         En-Zh & CNN, Arixv, Wikipedia, Quora& 225.6 & 1,800 \\
    \midrule
         De-En & Focus, Quora& 138.1 & 9,00 \\
    \bottomrule
    \end{tabular}
    \caption{Data statistics of document-level translations.}
    \label{tab:statistics}
\end{table}

The data statistics of the collected source documents are presented in Table~\ref{tab:statistics}.

\section{Translationese Span Annotation}
\label{app:guide}
\begin{CJK*}{UTF8}{gbsn}
Following the definition in Unbabel's guideline\footnote[1]{\url{https://help.unbabel.com/hc/en-us/articles/6444304419479-Annotation-Guidelines-Typology-3-0\#h\_01G4EYRD4K2KR9WKZ9WVT1N71K}}, in this work, we define translationese as too literal translations of the source. Through preliminary research, we generally categorized the issue into three subcategories: Unnatural Sentence Flow, Unnatural Phrase Flow, and Culture-specific Reference (e.g. Source: We don’t walk under ladders. Target: 我们不会在梯子下行走). Notably, the first two categories are more prevalent in LLM translation (see examples in Appendix~\ref{app:case}); therefore, this study focuses primarily on these two types.
\end{CJK*}
\par
We give our annotators a brief guideline and make detailed explanations with examples corresponding to each error category. 
Then, annotators are required to highlight all spans characterized as translationese errors in the document-level translation. 
During annotation, all translations of one given source are provided sequentially as a batch for the convenience of comparisons among different models (note that annotators do not know which model generated each translation, and the appearance order of translated documents is shuffled). 
The guideline for span annotation is shown as follows (see also Table \ref{style_MQM}):\\
\\
You will assess model translations of a source document, where each document may contain one or more sentences. Each target-language document is aligned with its corresponding source-language document, and both are displayed simultaneously on the annotation platform. For each model translation, identify and annotate spans with the specified error types. Annotate documents sequentially, as if reading them naturally. You may revisit and revise previously annotated documents as needed.

\begin{enumerate}
    \item {The key issues in this task are style errors and unnatural expressions (so-called translationese). You can label one expression as long as it seems to be strange from the perspective of the contemporary target language. To identify an error, highlight the relevant span of text, and select a category from the available options.}
    \item {When identifying errors, please identify all errors within each translated document and be as fine-grained as possible. For example, if there are two separate unnatural phrases in one sentence, please annotate two phrases respectively instead of selecting the whole sentence.}
    \item {Besides the three categories of style errors we provided, there are also some categories of translation errors for mistranslation situations. If it is not possible to reliably identify distinct errors because the translation is too badly garbled or is unrelated to the source, then mark a single Non\-translation error that spans the entire document.}
\end{enumerate}

\section{Annotation Implementation}
\begin{figure*}[!t]
    \centering

    \begin{subfigure}[b]{\textwidth}  
        \centering
        \includegraphics[width=\textwidth]{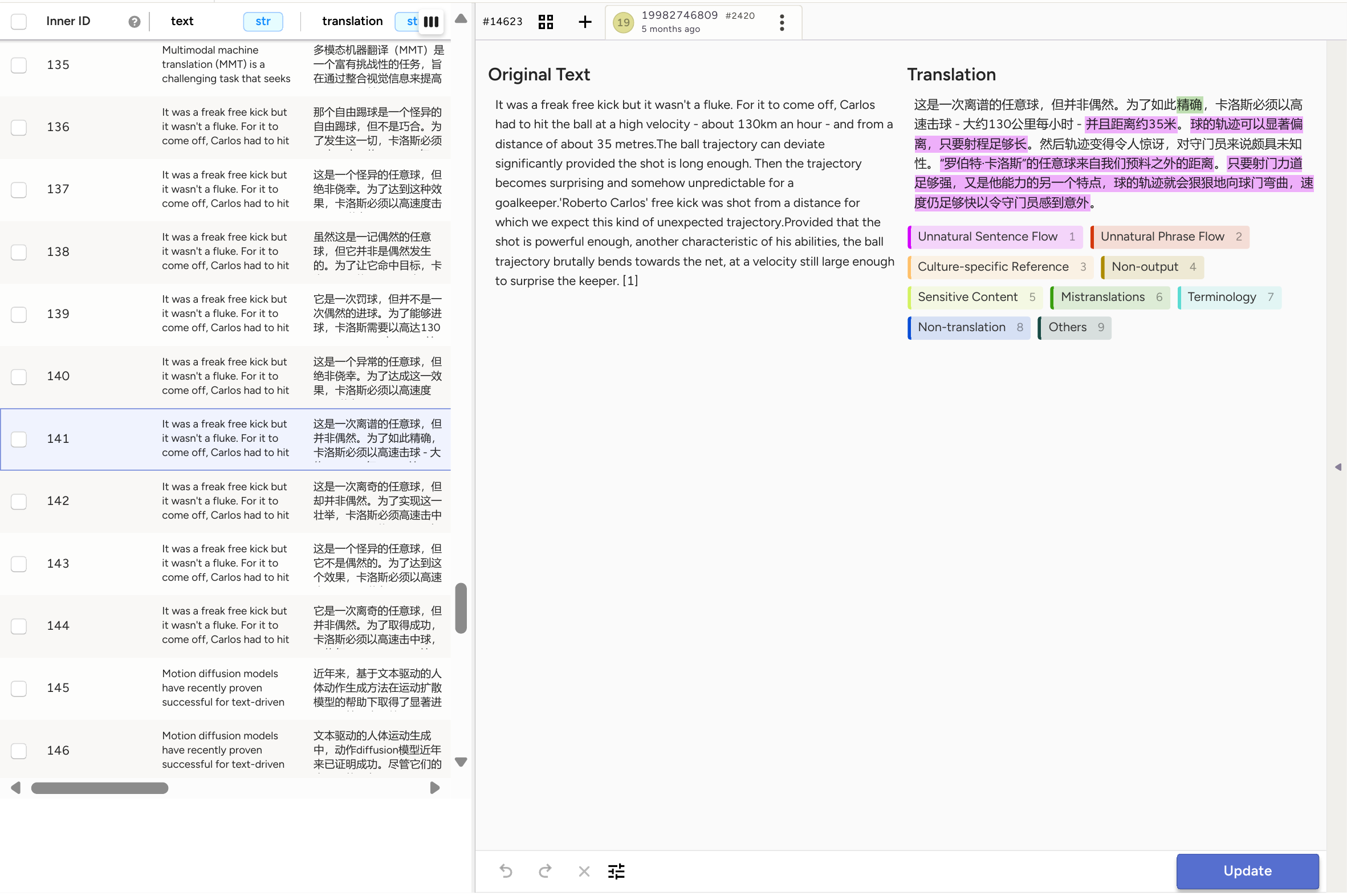}  
        \label{fig:label_studio1}
    \end{subfigure}
    
    \vspace{-1em}  

    \begin{subfigure}[b]{\textwidth}  
        \centering
        \includegraphics[width=\textwidth]{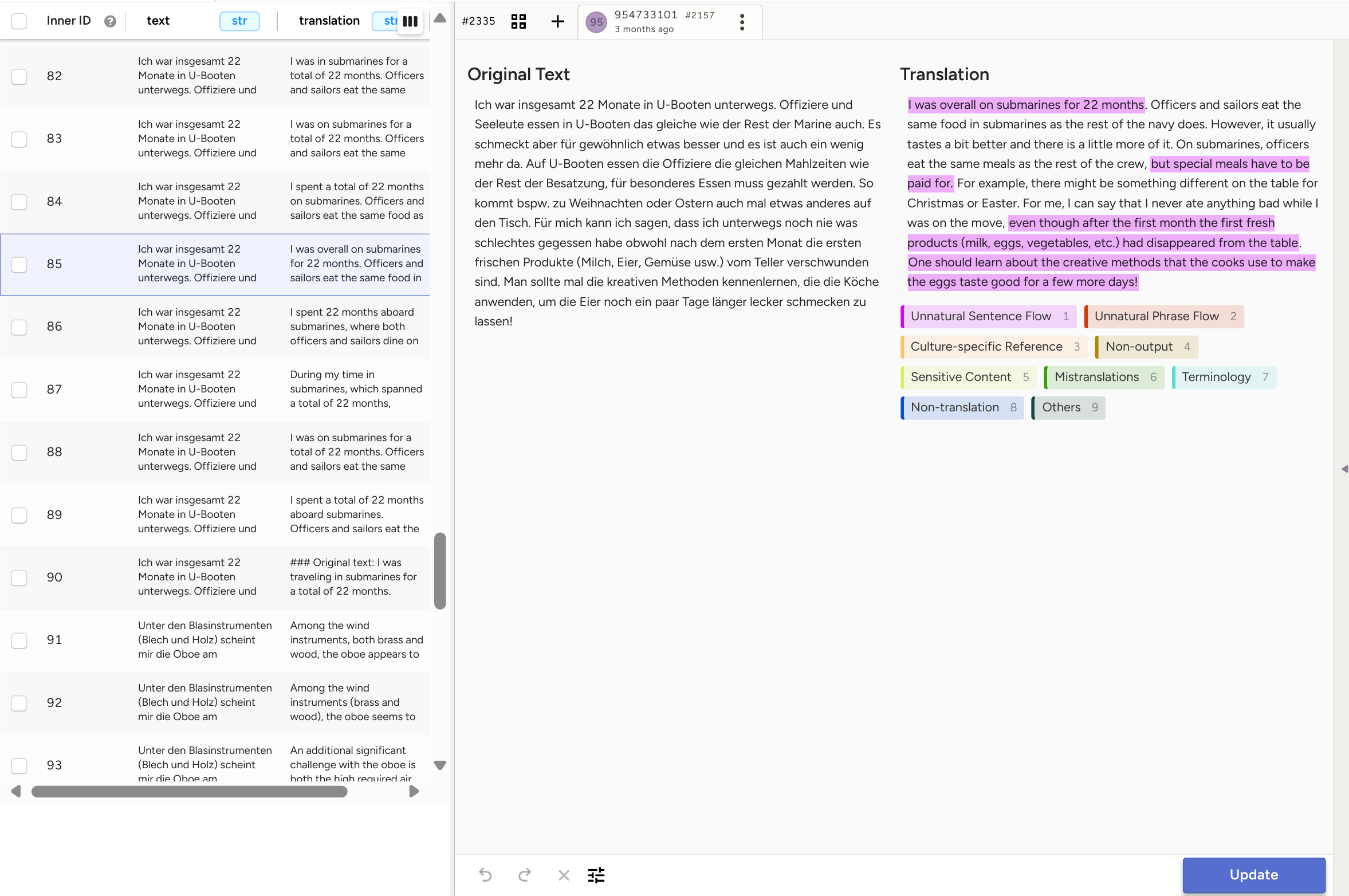}  
        \label{fig:label_studio2}
    \end{subfigure}

    \caption{Annotation platform demonstration (English-Chinese and German-English).}
    \label{fig:label_studio}
\end{figure*}

Based on the above guideline, we develop a specialized annotation platform using Label Studio~\cite{label_studio}, as demonstrated in Figure~\ref{fig:label_studio}.\par
The annotation tasks are conducted in batches, with each batch containing 180 translated documents corresponding to 20 source texts. As mentioned above, translations generated by different models from the same source text are presented simultaneously, but in a randomized order. Given the potential subjectivity in annotators' judgments on translationese, the results of annotation are subsequently reviewed by a senior annotator. This process aims to prevent significant disparities in annotating standards. Each batch of annotations takes approximately 16 hours for English-Chinese direction and 24 hours for German-English. The total time cost is 160 hours and 120 hours, respectively.

\begin{table}[t!]
    \small
    \centering
    \begin{tabular}{cccc}
    \toprule
    \multicolumn{4}{c}{English-Chinese Translation}  \\
    \midrule
        \textbf{Judge} &  \textbf{A-1} & \textbf{A-2} & \textbf{A-3}  \\
    \midrule
         \textbf{A-1} &  -& 0.592 & 0.742  \\
    \midrule
         \textbf{A-2}  & 0.592 & - & 0.603 \\
    \midrule
         \textbf{A-3} & 0.742 & 0.603 &  - \\
    \midrule
    \multicolumn{4}{c}{German-English Translation}  \\
    \midrule
        \textbf{Judge} &  \textbf{A-1} & \textbf{A-2} & \textbf{A-3}  \\
    \midrule
         \textbf{A-1} &  -& 0.753 & 0.587  \\
    \midrule
         \textbf{A-2}  & 0.753 & - & 0.553 \\
    \midrule
         \textbf{A-3} & 0.587 & 0.553 &  - \\
    \bottomrule
    \end{tabular}
    \caption{Inter-annotator agreement (Kendall's Tau scores) on naturalness voting.}
    \label{tab:iaa}
\end{table}


\section{TSR Scores}
\label{app:tsr}
The evaluation of the translationese span ratio for all models under both translation directions is presented in Table~\ref{tab:tsr}.

\begin{table*}[t]
\centering
\small
\begin{tabular}{lccccccccc}
\toprule
 \multirow{2}{*}{\textbf{Direction}} & \multirow{2}{*}{ALMA-7B} & \multirow{2}{*}{ALMA-13B} & \multirow{2}{*}{Mistral-7B} & \multicolumn{3}{c}{GPT-3.5} & \multicolumn{3}{c}{GPT-4}\\
  & & & & Direct & Specified & Polishing & Direct &  Specified & Polishing \\ 
  \midrule
 \textbf{En-Zh} & 0.19   & 0.18 & 0.32 & 0.22 & 0.23  & 0.20   & 0.20 & 0.17   & \textbf{0.14} \\ 
 \midrule
 \textbf{De-En} & 0.23  & 0.23  & 0.22  & 0.21 & 0.22  & 0.20 & 0.21 & 0.21  & \textbf{0.19}  \\ 
 \bottomrule
\end{tabular}
\caption{Translationese span ratios of different LLMs in English-Chinese and German-English translations.}
\label{tab:tsr}
\end{table*}

\section{Case Study of Translationese}
\label{app:case}
We demonstrate several real translation cases of both translationese errors in Table~\ref{table:case:en-zh} (English-Chinese) and Table~\ref{table:case:de-en} (German-English).

\section{Sentence-level Annotation}
\label{app:guide_sent}
Annotators are assigned another translation assessment task at the sentence level. They are required to follow the same guideline shown in Appendix~\ref{app:guide} as well. Similarly, each sentence is aligned with a corresponding source sentence. Annotators are asked to read in sequential order, with permission to revise previous sentences. The total time cost is 16 hours (English-Chinese) and 24 hours (German-English), respectively.

\section{Training Details}
\label{app:train}
All models are fine-tuned using LoRA~\cite{lora} with a rank of 16, employing a batch size of 16 on an A100 GPU.
The learning rate is set to \(1 \times 10^{-4}\) with a warmup ratio of 0.1.
Training is conducted for three epochs, selecting the model that achieves the lowest validation loss.
We perform training using Llama-Factory~\cite{factory} and leverage Deepspeed~\cite{ds} to accelerate training.

\section{Human Ranking}
\label{app:human_rank}
In the voting task, annotators are given a file in which each source document is aligned with three distinctive translations. They are required to rank the severity of translationese issues in each translation. A higher rank indicates less translationese and more natural language flow. When making judgments about translationese. Annotators still follow the guideline we provided for span annotation, but we do not provide a specific breakdown of the ranking scheme. The total time cost is 24 hours (English-Chinese) and 32 hours (German-English), respectively.
The inter-annotator agreement evaluation is presented in Table~\ref{tab:iaa}.

\section{Case Study of SFT Methods}
\label{app:case2}
Cases of translations from SFT, SFT-KD and STF-Polished are also demonstrated in Table~\ref{table:case_polish:en-zh} (English-Chinese) and Table~\ref{table:case_polish:de-en} (German-English).

\begin{table*}[h]
\centering
\small
\resizebox{1.0\linewidth}{!}{
\begin{tabular}{p{3.7cm} p{7cm}}
\toprule
\textbf{Error Category} & \textbf{Description}\\
\midrule
\textbf{Unnatural Sentence Flow} & {A sentence-level translation issue where the structure of the sentence is considered unnatural in the target language. This often occurs when complex sentence structures from the source language are directly translated, resulting in sentences that are difficult to read in the target language.}\\
\midrule
\textbf{Unnatural Phrase Flow} & {A portion of text, larger than a single word or multiword expression, is a too literal translation of the source. The meaning of the source comes through in the target, but the overall feeling of the translation is unnatural. }\\
\midrule
\textbf{Culture-specific Reference} & {The target text contains a culture-specific reference that’s not appropriate or understandable to the intended target audience. An example of this is the use of jargon related to sports or other culture-specific features that are not necessarily understood in the environment of the target language.}\\
\midrule
\textbf{Sensitive Content} & {The presence of sensitive information in the translation or source text, such as references to violence, war, etc.}\\
\midrule
\textbf{Mistranslation} & {Minor errors including mistranslations, omissions, or over-translations.}\\
\midrule
\textbf{Terminology} & {Errors related to the incorrect use of domain-specific terms or technical jargon.}\\
\midrule
\textbf{Non-translation} & {Impossible to reliably characterize distinct errors (or the model repeatedly outputs meaningless contents)}\\
\midrule
\textbf{Others} & {Errors that affect the readability and naturalness of the text but do not fit neatly into the other defined categories. Annotators should provide specific comments on these errors.}\\
\bottomrule
\end{tabular}
}
\caption{
\label{style_MQM}
\centering
Annotation Guideline in the present study
}
\end{table*}


    



\begin{table*}[t]
\centering
\small
\begin{CJK*}{UTF8}{gbsn}
\resizebox{1.0\linewidth}{!}{
\renewcommand{\arraystretch}{1.15}
\begin{tabular}{p{4cm}p{1.5cm}p{7cm}}
\toprule
\textbf{Error Category} & \multicolumn{2}{c}{\textbf{Example}}\\
\midrule
\multirow{16}*{\textbf{Unnatural Sentence Flow}} & \textbf{Source} & {Our benchmarking findings can serve future research aiming to improve the generic capability of LMs on semantic phrase comprehension.}\\
& \textbf{Translation} & {我们的评测结果将为未来研究，旨在提升语言模型在语义表达理解任务中的普适能力，提供有价值的参考。}\\
\cline{2-3}
& \textbf{Source} & {An analysis of a core cohort comprising 380 articles from multiple disciplines captures the most recent advancements in responsible AI.}\\
& \textbf{Translation} & {通过一个包括来自多个学科的380篇文章的核心队列的分析，捕捉了负责任AI的最新进展。}\\
\cline{2-3}
& \textbf{Source} & {They both contribute to the development of a unified model that is highly generalizable, versatile, and comprehensible for time series analysis.}\\
& \textbf{Translation} & {二者共同促进了高度通用、多功能且易于理解的统一模型的发展，用于时间序列分析。}\\
\midrule
\multirow{6}*{\textbf{Unnatural Phrase Flow}} & \textbf{Source} & {demonstrated remarkable improvements}\\
& \textbf{Translation} & {展示了显著的改进}\\
\cline{2-3}
& \textbf{Source} & {demonstrating promising performance}\\
& \textbf{Translation} & {展示了有希望的性能}\\
\cline{2-3}
& \textbf{Source} & {credit risk management is particularly core}\\
& \textbf{Translation} & {信用风险管理尤为核心}\\
\bottomrule
\end{tabular}
}
\caption{
\label{table:case:en-zh}
\centering
Samples of translationese errors in large language model translation (English-Chinese).
}
\end{CJK*}
\end{table*}

\begin{table*}[t]
\centering
\small
\resizebox{1.0\linewidth}{!}{
\renewcommand{\arraystretch}{1.15}
\begin{tabular}{p{4cm}p{1.5cm}p{7cm}}
\toprule
\textbf{Error Category} & \multicolumn{2}{c}{\textbf{Example}}\\
\midrule
\multirow{13}*{\textbf{Unnatural Sentence Flow}} & \textbf{Source} & {So geht es nicht, findet die italienische Regierung und ließ Dutzende von elektrischen Fiat Topolinos beschlagnahmen.}\\
& \textbf{Translation} & {This is not acceptable, finds the Italian government and seized dozens of electric Fiat Topolinos.}\\
\cline{2-3}
& \textbf{Source} & {Das zweite Gruppenspiel bestreitet die DFB-Elf fünf Tage später am 19. Juni in Stuttgart gegen Ungarn.}\\
& \textbf{Translation} & {The second group game will be played five days later on 19 June in Stuttgart against Hungary.}\\
\cline{2-3}
& \textbf{Source} & {Nach meinem Wissen sind wir die Ersten in Deutschland, die das angewendet haben, sogar in Europa}\\
& \textbf{Translation} & {To the best of my knowledge, we are the pioneers in Germany in using it, even in Europe,}\\
\midrule
\multirow{6}*{\textbf{Unnatural Phrase Flow}} & \textbf{Source} & {schufen aber einen rockigeren sound}\\
& \textbf{Translation} & {crafted a grittier sound}\\
\cline{2-3}
& \textbf{Source} & {sie sich stark mit anderen Arten vermischt}\\
& \textbf{Translation} & {it mixes strongly with other species}\\
\cline{2-3}
& \textbf{Source} & {sind wir in der USA in Urlaub gewesen}\\
& \textbf{Translation} & {we vacationed in the US}\\
\bottomrule
\end{tabular}
}
\caption{
\label{table:case:de-en}
\centering
Samples of translationese errors in large language model translation (German-English).
}
\end{table*}

\begin{table*}[t]
\centering
\small
\begin{CJK*}{UTF8}{gbsn}
\resizebox{1.0\linewidth}{!}{
\renewcommand{\arraystretch}{1.15}
\begin{tabular}{p{2cm}p{10cm}}
\toprule
\multicolumn{2}{c}{\textbf{Example}}\\
\midrule
\textbf{Source} & {50 countries had purchased rights to the telecast, \textcolor{blue}{and when the numbers were added up, it was estimated that 300 million people worldwide had watched the fight}, including 27.5 million people in the UK, about half the national population at the time. \textcolor{blue}{To put into perspective how big the fight was just in the UK}, more people watched the fight than the closing ceremony of the 2012 Olympics games (which was in London) and the full coverage of the 2011 Royal wedding of Prince William and Kate (a British royal wedding), despite neither Frazier or Ali originating from the UK, and the fight taking place in America.}\\
\cline{1-2}
\textbf{SFT} & {50 个国家购买了转播权，\textcolor{red}{当这些数字相加时，估计有 3 亿人次在全世界观看了这场比赛}，包括 2750 万英国观众，这几乎是当时英国人口的一半。\textcolor{red}{为了将这场比赛在英国的规模置于合适的位置}，我们可以看到，观众人数比 2012 年伦敦奥运会闭幕式和 2011 年威廉王子和凯特的婚礼（一场英国皇室婚礼）更高，尽管弗雷泽和阿里都不是英国人，比赛也在美国举行。}\\
\textbf{SFT-KD} & {50个国家购买了广播权，\textcolor{red}{当数字相加时，全球估计有3亿人观看了这场比赛}，其中包括英国的2750万人，这约占当时国民总数的一半。\textcolor{red}{为了让大家对英国本地观看人数有一个感性认识}，观看这场比赛的人数超过了2012年伦敦奥运会闭幕式和2011年威廉王子与凯特的皇家婚礼的全覆盖（尽管弗雷泽和阿里都不是来自英国，而且这场比赛是在美国举行的）。}\\
\textbf{SFT-Polished} & {50个国家已经购买了电视转播权。\textcolor{red}{加总后，全球观众估计超过了3亿人次}，其中包括英国的2750万人，约占当时英国国民总数的一半。\textcolor{red}{为了更准确地理解这一拳赛在英国的影响力}，尽管弗雷泽和阿里均非英国人，并且该比赛是在美国举行的，观众人数仍然超过了2012年伦敦奥运会闭幕式和2011年威廉王子与凯特的联合王国皇室婚礼的观众人数。}\\
\midrule
\textbf{Source} & {\textcolor{blue}{Rice's biggest success was her first novel}, "Interview with the Vampire," which was published in 1976 and introduced the character of the vampire Lestat, \textcolor{blue}{who would be the central character in the 13-book Chronicles series}, the most recent of which was published in 2018.}\\
\cline{1-2}
\textbf{SFT} & {\textcolor{red}{米勒的最大成就在于她的第一部小说}《吸血鬼采访》，该小说于 1976 年出版，引入了吸血鬼莱斯特这个角色，\textcolor{red}{这位角色将成为 13 部系列小说中的主角}，这些系列小说中的最新一部于 2018 年出版。}\\
\textbf{SFT-KD} & {\textcolor{red}{米尔斯最大的成功是她的第一部小说}《吸血鬼的采访》，该小说于1976年出版，引入了吸血鬼莱斯特的角色，\textcolor{red}{这位角色将成为13部书籍系列《纪事》的核心人物}，该系列的最新作品于2018年出版。}\\
\textbf{SFT-Polished} & {\textcolor{red}{米尔斯最大的成就是她的首部小说}《吸血鬼之访谈》，该书于1976年出版，首次引入了吸血鬼莱斯特特的角色，\textcolor{red}{这位角色后来成为了《吸血鬼纪事》系列中的核心人物，该系列共13部}，最新的一部在2018年出版。}\\
\bottomrule
\end{tabular}
}
\caption{
\label{table:case_polish:en-zh}
\centering
Samples of translations from SFT, SFT-KD and SFT-Polished (English-Chinese).
}
\end{CJK*}
\end{table*}

\begin{table*}[t]
\centering
\small
\begin{CJK*}{UTF8}{gbsn}
\resizebox{1.0\linewidth}{!}{
\renewcommand{\arraystretch}{1.15}
\begin{tabular}{p{2cm}p{10cm}}
\toprule
\multicolumn{2}{c}{\textbf{Example}}\\
\midrule
\textbf{Source} & {Trockener Rasen und gekieste Wege mögen für andere heimische Nacktschnecken ein Problem sein, nicht aber für Arion vulgaris. \textcolor{blue}{Sie vermehrt sich Experten zufolge schneller}, frisst mehr und setzt sich notfalls zum Fressen in die pralle Sonne, ohne Schaden zu nehmen. Zudem zeigen Erbgutanalysen, dass \textcolor{blue}{sie sich stark mit anderen Arten vermischt} - und sich auf diese Weise womöglich immer neue günstige Eigenarten für die jeweilige Umgebung aneignet. Und als wäre das alles nicht genug: \textcolor{blue}{An einer ausgewachsenen Arion vulgaris haben - von Indischen Laufenten abgesehen - kaum Fressfeinde Interesse}. Auch manche Laufkäfer können junge Wegschnecken oder deren Eier fressen.}\\
\cline{1-2}
\textbf{SFT} & {Dry lawn and cobbled paths may be a problem for other native slug species, but not for Arion vulgaris. \textcolor{red}{Experts say that it multiplies faster}, eats more, and if need be, sets itself down in the full sun to eat, without suffering any damage. In addition, DNA analyses show that \textcolor{red}{it mixes strongly with other species} - and thus acquires new favorable characteristics for the respective environment. And if that were not enough: \textcolor{red}{Only Indian runners have any interest in eating an adult Arion vulgaris.} Even some beetle larvae can eat young slugs or their eggs.}\\
\textbf{SFT-KD} & {Dry lawns and graveled paths may be a problem for other native slugs, but not for Arion vulgaris. \textcolor{red}{According to experts, it reproduces faster}, eats more, and if necessary, basks in the full sun to feed, without any harm. Moreover, genetic analyses show that \textcolor{red}{it strongly hybridizes with other species} - and in this way, possibly acquires ever new advantageous characteristics for its environment. And if all that were not enough: \textcolor{red}{Among fully grown Arion vulgaris, hardly any predators have an interest - except for Indian leaf turtles.} Some ground beetles can also eat young slugs or their eggs.}\\
\textbf{SFT-Polished} & {Dry lawns and gravel paths may pose challenges for other native slugs, but not for Arion vulgaris. \textcolor{red}{Experts note that it reproduces more rapidly}, consumes more, and, if necessary, basks in the full sun without suffering any harm. Additionally, genetic analyses reveal that \textcolor{red}{it has extensively interbred with other species}, potentially acquiring new adaptive traits for its specific environments. Moreover, \textcolor{red}{apart from the Indian land tortoise, few predators show interest in an adult Arion vulgaris.} Some ground beetles, however, can consume young slugs or their eggs.}\\
\midrule
\textbf{Source} & {"Tatsächlich haben meine Frau und ich 3 Dosen des Biontech-Pfizer-Impfstoffs erhalten", \textcolor{blue}{schrieb Şahin am Freitag in einem Beitrag} im Netzwerk "LinkedIn"}\\
\cline{1-2}
\textbf{SFT} & {"In fact, my wife and I have received 3 doses of the Biontech/Pfizer vaccine", Şahin wrote \textcolor{red}{on Friday in a contribution to} the LinkedIn network.}\\
\textbf{SFT-KD} & {Indeed, my wife and I have received 3 doses of the BioNTech-Pfizer vaccine", Şahin wrote \textcolor{red}{on Friday in a post on} the "LinkedIn" network.}\\
\textbf{SFT-Polished} & {"Indeed, my wife and I have received three doses of the BioNTech-Pfizer vaccine", Şahin \textcolor{red}{wrote in a Friday post on} the LinkedIn network.}\\
\bottomrule
\end{tabular}
}
\caption{
\label{table:case_polish:de-en}
\centering
Samples of translations from SFT, SFT-KD and SFT-Polished (German-English).
}
\end{CJK*}
\end{table*}

\end{document}